\documentclass[letterpaper, 10 pt, conference]{ieeeconf}  

\IEEEoverridecommandlockouts                              

\usepackage{xspace}
\newcommand{\smriti}{ReCAT\xspace}

\usepackage{cite}
\usepackage{microtype}

\usepackage{tabularx}
\usepackage{booktabs}
\usepackage[table]{xcolor}
\definecolor{oursblue}{RGB}{234,241,250}
\definecolor{liberogray}{RGB}{242,244,247}

\newcommand{\tbd}{\textcolor{gray}{--}}

\usepackage{pgfplots}
\pgfplotsset{compat=1.18}
\usepgfplotslibrary{groupplots}
\usetikzlibrary{calc}
\definecolor{cbBlue}{HTML}{0072B2}
\definecolor{cbOrange}{HTML}{E69F00}
\definecolor{cbGreen}{HTML}{009E73}
\definecolor{cbRed}{HTML}{D55E00}
\definecolor{cbSky}{HTML}{56B4E9}
\definecolor{capA}{HTML}{C6DBEF}
\definecolor{capB}{HTML}{6BAED6}
\definecolor{capC}{HTML}{2171B5}
\definecolor{capD}{HTML}{08306B}

\pgfplotsset{
  paperaxis/.style={
    width=\columnwidth, height=4.3cm,
    ymin=0, ymax=100, ytick distance=25,
    ylabel={Success (\%)},
    grid=major, grid style={line width=.2pt, draw=gray!22},
    axis line style={line width=.4pt, draw=gray!55},
    tick style={line width=.4pt, draw=gray!55},
    tick label style={font=\scriptsize},
    label style={font=\scriptsize},
    legend style={font=\scriptsize, draw=none, fill=none,
                  at={(0.5,1.02)}, anchor=south, legend columns=-1,
                  /tikz/every even column/.append style={column sep=6pt}},
  },
  paperbar/.style={paperaxis, ybar, bar width=4.5pt,
                   legend style={font=\scriptsize, draw=none, fill=none,
                                 at={(0.5,1.02)}, anchor=south, legend columns=2,
                                 /tikz/every even column/.append style={column sep=5pt}},
                   enlarge x limits=0.22, xtick=data,
                   xticklabel style={font=\scriptsize}},
}

\definecolor{cPurple}{HTML}{9370DB}
\definecolor{oursLight}{HTML}{D9822B}
\definecolor{oursMed}{HTML}{A66300}
\definecolor{oursDark}{HTML}{7A4800}

\pgfplotsset{
  lineaxis/.style={
    width=0.36\textwidth, height=4cm,
    ymin=0, ymax=100, ytick={0,25,50,75,100},
    grid=major, grid style={line width=.2pt, draw=gray!20},
    axis line style={line width=.4pt, draw=gray!55},
    tick style={line width=.4pt, draw=gray!55},
    tick label style={font=\scriptsize},
    label style={font=\scriptsize},
    title style={font=\scriptsize},
    enlarge x limits=0.15,
    legend style={font=\scriptsize, draw=none, fill=none,
                  at={(0.5,1.05)}, anchor=south, legend columns=-1,
                  /tikz/every even column/.append style={column sep=5pt}},
    every axis plot/.append style={line width=1pt, mark size=2.2pt},
  },
}
\newcommand{\finalstageband}{%
  \fill[gray!10] (rel axis cs:0.70,0) rectangle (rel axis cs:1,1);
}

\usepackage{graphicx}
\usepackage{placeins}  
\usepackage{caption}  
\usepackage{amsmath}
\usepackage{amssymb}

\title{\LARGE \bf
ReCAT: Remember, Count, and Time: Structured Recurrent Memory for Robot Manipulation 
}

\author{%
  Pankhuri Vanjani$^{1*}$, Mostafa Hatab$^{1*}$, Can Mizrakli$^{1*}$, Vaisakh Shaj$^{2}$,
  Zhuoyue Li$^{1}$,\\ Moritz Reuss$^{3}$, Rudolf Lioutikov$^{1,4}$\\[3pt]
  \normalsize $^{1}$Intuitive Robots Lab, Karlsruhe Institute of Technology (KIT), Germany \quad
               $^{2}$University of Edinburgh, UK\\
  \normalsize $^{3}$NVIDIA \quad
               $^{4}$Robotics Institute Germany (RIG)\\[3pt]
  \quad
  \normalsize %
  \thanks{ \normalsize $^{*}$Equal contribution}%
}

\begin{document}

\maketitle
\thispagestyle{empty}
\pagestyle{empty}

\begin{abstract}
Memory-dependent manipulation requires robots to make decisions using information that is no longer available to their current sensors, such as recalling an earlier visual cue, tracking task progress, counting repeated events, or estimating elapsed time. We present \smriti, a language-conditioned policy with structured recurrent memory. An instruction-conditioned encoder forms features from the current observation. A recurrent memory integrates the observation stream through Mamba-2 layers and one causal attention layer. A flow-matching Transformer decoder reads the current and the historical representation through separate cross-attention in every block. \smriti reaches 95.3\% average success on LIBERO and 62.4\% on RMBench, with the best or tied-best result on six of nine tasks. On three real-robot tasks probing spatial recall, event counting, and interval timing, the best \smriti variant reaches 66.7\% average success, against 8.3\% for the strongest short-history baseline. Controlled comparisons within \smriti show that the observation encoder and every-block memory conditioning are needed for this performance. They also show that update rules developed for efficient sequence modeling behave differently as robot memory: additive updates have the highest observed success on counting and timing, and delta-rule updates on spatial recall. Project website is at https://intuitive-robots.github.io/ReCAT

\end{abstract}


\section{INTRODUCTION}

\begin{figure}[t]
    \centering
    \includegraphics[width=\columnwidth]{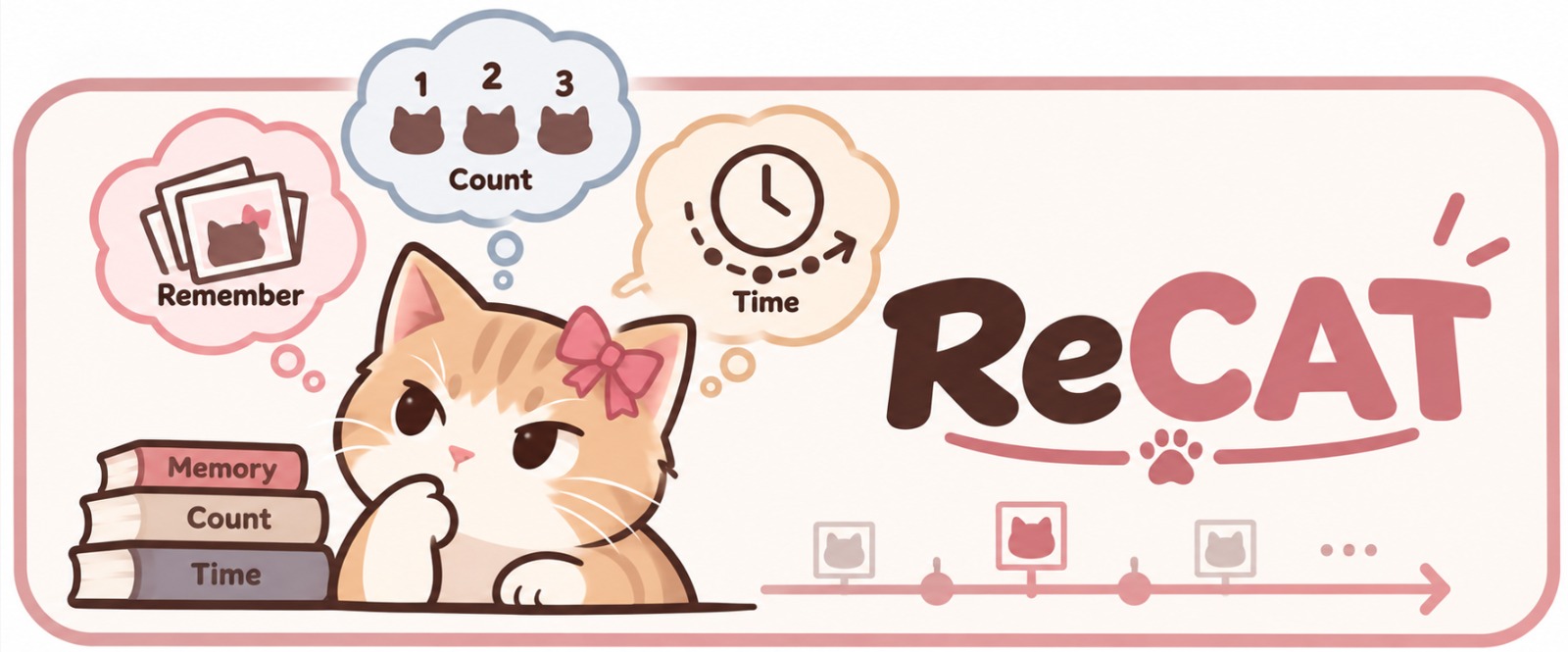}\\[3pt]
    \includegraphics[width=\columnwidth]{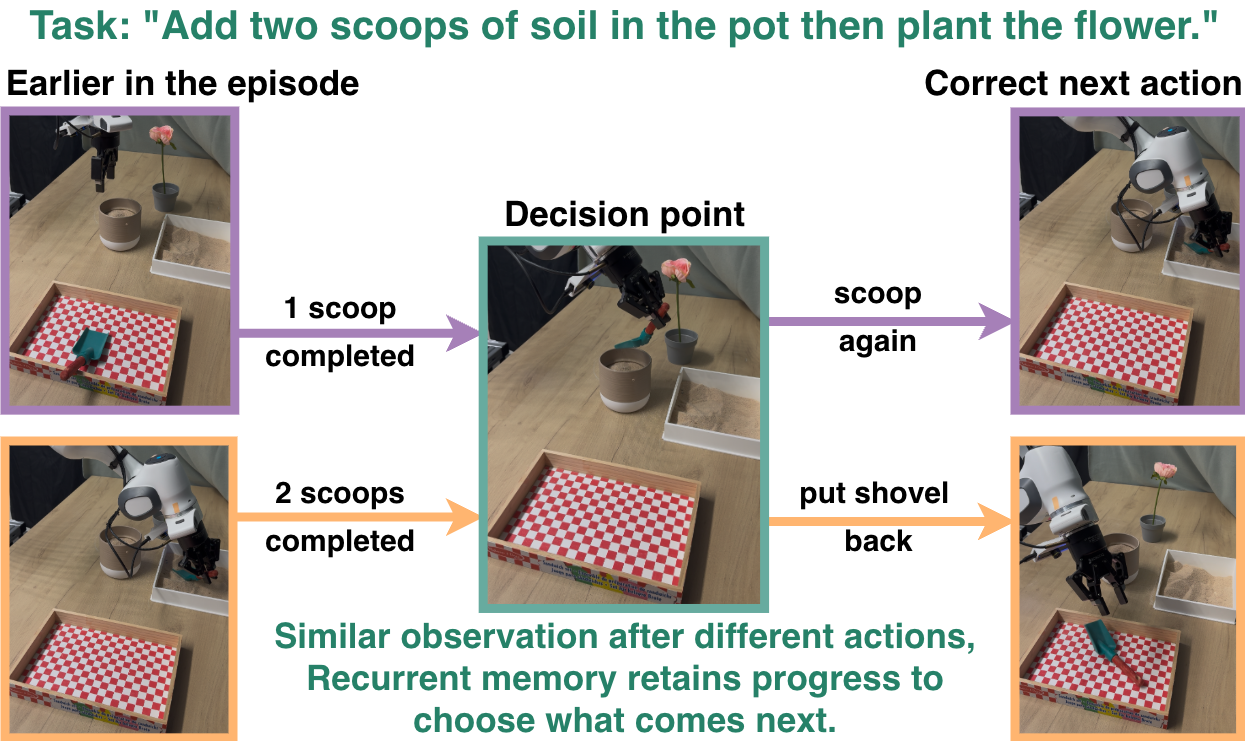}
    \caption{Visually similar observations can require different actions depending on task history. \smriti retains the completed scoop count in a compact recurrent memory, allowing the robot to scoop again after one scoop and put down the shovel after two.}
    \label{fig:teaser}
\end{figure}

Robot manipulation often requires information that is no longer available in the current observation. An object's original location may have left the camera view, the number of completed repetitions may not be apparent, or the next action may depend on how much time has elapsed. In the \textsc{Plant} task (Fig.~\ref{fig:teaser}), for example, the shovel-over-pot view looks alike after one scoop and after two, yet the required actions differ: scoop again, or put down the shovel and proceed to planting. Such tasks are non-Markovian with respect to the current observation, and the policy must combine current perceptual information with relevant interaction.

Policies obtain this history in different ways. Some use a window of recent observations. Some store selected observations in a memory bank~\cite{shi2026memoryvla,li2026remem}. Others summarize the whole observation stream in a compact recurrent state~\cite{gu2021efficiently,yang2024gated,lahoti2026mamba, vanjani2026dam}. Recent and concurrent work shows that such full-history recurrence helps in manipulation~\cite{zhou2025mtil,yoo2025robossm,tsuji2025mamba,guan2026dssp,zhou2026chronos,cherepanov2026mu}. Yet, precise manipulation also requires the current spatial and proprioceptive observation. So, a policy needs both a history representation and direct access to the present.

We introduce \smriti, a structured memory policy for language-conditioned manipulation. An instruction-conditioned encoder turns each observation into multimodal features. These features take two paths. One path feeds the action decoder directly. The other path updates a temporal memory. This memory is a stack of recurrent layers with one causal attention layer in the middle. A flow-matching Transformer decoder reads the current and the historical representation through separate cross-attention in every block. \smriti therefore keeps a compact history representation and still sees the current observation directly when it generates actions.

We evaluate \smriti on LIBERO, RMBench, and three real-robot tasks. The real-robot tasks probe spatial recall, event counting, and interval timing. \smriti reaches 95.3\% average success on LIBERO and 62.4\% on RMBench. On the real-robot tasks it reaches 66.7\%, compared with 8.3\% for the strongest short-history baseline. Stage-wise evaluation shows where a policy fails: during manipulation, or when it must use history. Controlled comparisons within \smriti then show how observation encoding, recurrent update rule, memory capacity, and decoder conditioning affect performance. Mamba variants reach the highest success on counting and timing tasks.

\textbf{Our contributions are as follows: }

\begin{itemize}

\item We present \smriti, a structured policy whose temporal memory combines
recurrent layers with a single causal attention layer. It conditions
flow-matching action generation on separate current-observation and
history pathways.

\item \smriti reaches 95.3\% on LIBERO, 62.4\% on RMBench with the best or
tied-best result on six of nine tasks, and 66.7\% across three real-robot
memory tasks. It has the lowest measured inference latency among the
evaluated memory-based policies.

\item Controlled architectural and stage-wise comparisons show how
observation encoding, recurrent update rule, memory capacity, and decoder
conditioning affect immediate manipulation and later history-dependent
decisions.

\end{itemize}


\section{RELATED WORK}

\subsection{Memory in Visuomotor Imitation Learning}

Robot policies differ in how they represent the past. Many keep past observations and attend to them. Fixed-window methods condition on recent visual tokens~\cite{mark2026bpp,yin2026simplememvla}. Memory-bank methods store observations over longer horizons and retrieve them when needed~\cite{shi2026memoryvla,torne2026mem,shah2026memory}.These methods give direct access to earlier frames, as a result their attention cost grows with the number of retained tokens unless they select or compress. The second way carries a compact learned state through time. ReMem-VLA and $\mu$VLA propagate recurrent queries or latent memory tokens~\cite{li2026remem,cherepanov2026mu}. Gated Memory Policy adds a learned retrieval gate~\cite{gao2026gated}.

These policies show that a compact state can carry task history. Each of them, however, pairs one memory mechanism with one way of feeding it to the policy. This leaves open a question: \emph{how does the recurrent update itself shape what is remembered, and how should that memory be combined with the current observation for action generation?} \smriti is built around both parts of it. Its memory is a stack of recurrent layers with one causal attention layer, and the update rule inside that stack is interchangeable. It's flow-matching decoder attends to the current observation and to the memory through separate cross-attention in every block. Section~\ref{sec:eval} evaluates this design and compares update rules, capacities within it.


\subsection{Recurrent Memory Architectures for Robot Manipulation}
State-space models (SSMs) and related recurrent architectures have been
used both as action-generation backbones and as history encoders in robot
learning. MaIL uses a Mamba encoder-decoder for behavior cloning, Mamba
Policy combines Mamba and attention within a 3D diffusion policy, and
DiSPo applies an SSM across discretized action scales
~\cite{jia2024mail,cao2025mamba,oh2024dispo}.  AnoleVLA uses causal Mamba recurrence to fuse proprioceptive, visual, and language features for action chunk prediction ~\cite{takagi2026anolevla}. For temporal conditioning,
the Mamba motion encoder extracts temporal features, MTIL processes full
trajectories with Mamba-2, RoboSSM processes long in-context
demonstrations, and Embodied-SlotSSM maintains object-centric memory~\cite{tsuji2025mamba,zhou2025mtil,yoo2025robossm}.
Concurrent work conditions on the full observation history as well. DSSP pairs a Mamba history representation with a dynamics-aware objective and hierarchical prefix conditioning for an SSM diffusion denoiser~\cite{guan2026dssp}. Chronos combines a selective historical state with a physics-informed action prior~\cite{zhou2026chronos}. RoboTTT adapts fast weights at test time with Gated DeltaNet recurrence~\cite{jiang2026robottt}. These policies each fix one recurrent backbone and one route from memory to action. \smriti differs on both counts. Its temporal memory mixes recurrent layers with causal attention, and its action decoder receives the memory and the current observation through separate cross-attention rather than as a prefix or a prior.

These approaches draw on several families of linear recurrence. All keep a fixed-size state at constant per-step cost~\cite{katharopoulos2020transformers,gu2023mamba,dao2024transformers}. Additive memories, such as linear attention and Mamba-2, accumulate key-value writes under a learned decay~\cite{katharopoulos2020transformers,schlag2021linear,dao2024transformers}. Delta-rule models erase and rewrite information along matching key directions~\cite{yang2024parallelizing,yang2025gated}. A third family adds richer state dynamics through negative eigenvalues, reflections, or complex-valued rotations~\cite{grazzi2025unlocking,lahoti2026mamba}. All of them can be written as $S_t=A_tS_{t-1}+b_t$. The structure of $A_t$ sets the memory semantics: accumulation, replacement, or rotation. Pure recurrence has a known limit, though. A fixed-size state struggles with exact recall of earlier tokens~\cite{arora2023zoology}. Efficient long-context language models therefore mix a few attention layers into the recurrent stack, and this hybrid design is now common~\cite{dao2024transformers,team2025kimi}. Mamba Policy brings the same idea to a robot policy backbone~\cite{cao2025mamba}. \smriti uses this hybrid design in two places. Its observation encoder mixes SSM and attention layers over the tokens of the current frame. Its temporal memory mixes them across time, with one causal attention layer among the recurrent layers, and the recurrent update rule within that stack is interchangeable. The two parts are meant to complement each other: the recurrent layers track task progress at constant cost, while the attention layer can look back at a specific earlier observation.

\section{METHODOLOGY}

\begin{figure*}[t]
    \centering
    \includegraphics[width=\textwidth]{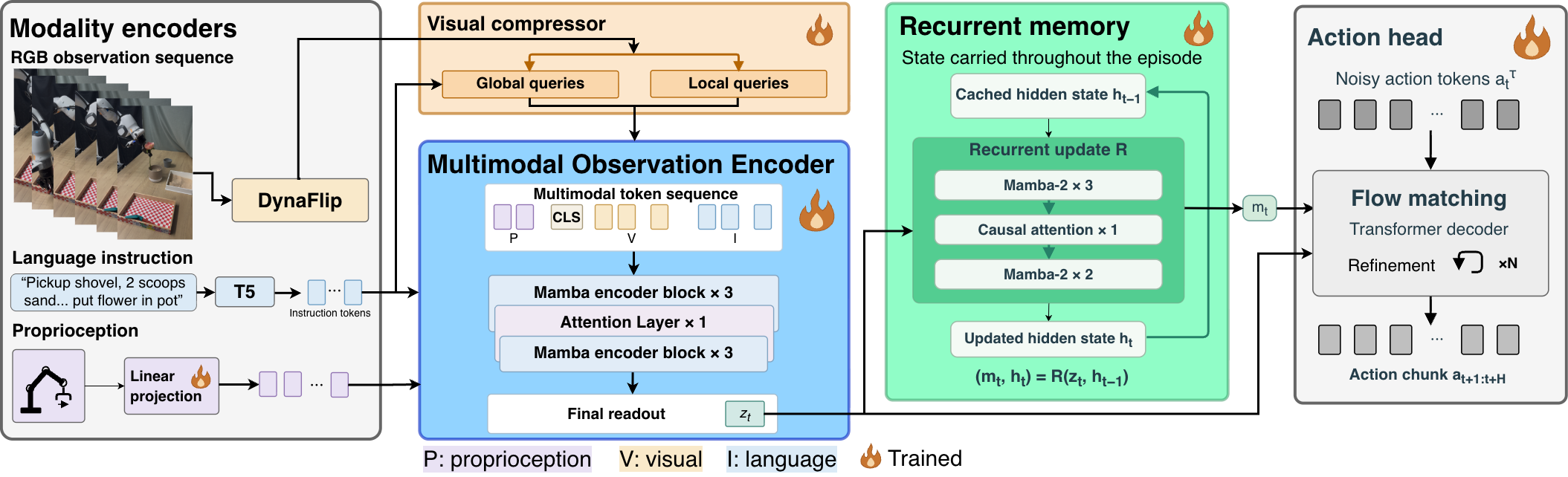}
    \caption{Overview of \smriti. Frozen DynaFLIP and T5 encoders (LoRA-adapted) feed a language-conditioned visual compressor and a six-layer Mamba--attention frame encoder, which produces one feature vector $z_t$ per step. The same $z_t$ feeds the action decoder directly and is the input to the recurrent memory, which returns the readout $m_t$. The flow-matching decoder attends to $z_t$ and $m_t$ through separate cross-attention in every block. The decoder backbone is shared across all experiments. The ablations vary the frame-encoder mixer, the recurrent update rule, memory depth and width, and how $m_t$ enters the decoder.}
    \label{fig:demo_architecture}
\end{figure*}

\subsection{Problem Formulation}

We consider language-conditioned imitation learning for partially observed robot manipulation. The training set consists of demonstrations $\mathcal{D}={(\ell^{(i)},(s_t^{(i)},a_t^{(i)}){t=1}^{T_i})}{i=1}^{N}$, where $\ell^{(i)}$ is the task instruction, $s_t^{(i)}=(v_t^{(i)},p_t^{(i)})$ combines the camera observations and proprioception, and $a_t^{(i)}\in\mathbb{R}^{d_a}$ is the demonstrated action. We denote the available observation history by $\mathcal{H}_t=(s_{1:t},\ell)$ and an action chunk of length $K$ by $A_t=(a_t,\ldots,a_{t+K-1})$.



In memory-dependent tasks, the current observation $s_t$ may be insufficient to determine the task state, so visually similar observations can require different actions depending on prior events. The policy is therefore non-Markovian with respect to $s_t$ alone and is modeled as $\pi_\theta(A_t\mid\mathcal{H}_t)$. Rather than storing the growing history $\mathcal{H}_t$, \smriti updates a compact recurrent representation as new observations arrive.
\vspace{-2mm}
\subsection{\smriti}
\smriti has three parts (Fig.~\ref{fig:demo_architecture}). An instruction-conditioned observation encoder turns the current images, instruction, and proprioception into one feature vector $z_t$. A temporal memory takes $z_t$ as its input at every step, updates its retained state $h_t$, and returns a history-conditioned readout $m_t$. A flow-matching action decoder predicts the next action chunk from $z_t$ and $m_t$. The same $z_t$ therefore feeds the decoder directly and drives the memory.

\paragraph{Current-Observation Encoding}

DynaFLIP~\cite{lee2026dynaflip} encodes each camera image into a $16\times16$ patch grid $P_t$. Its T5 text encoder produces instruction tokens $L$. Both backbones stay frozen and are adapted with LoRA. Following Compressor-VLA~\cite{gao2025compressor}, we compress each patch grid with language-conditioned global queries and local window queries. The global queries keep task-relevant context. The local queries keep fine spatial detail.
\begin{equation}
    Z_t=\mathrm{CrossAttn}
    \bigl(\gamma(\bar{\ell})\odot Q+\beta(\bar{\ell}),\,P_t\bigr),
    \label{eq:compressor}
\end{equation}
Here $\bar{\ell}$ summarizes the instruction, $Q$ are the learned global queries, and $\gamma,\beta$ are FiLM parameters. The local window queries work the same way over their windows. Global and local outputs are concatenated into 80 tokens per camera. A six-layer encoder then mixes these tokens with the instruction tokens and a proprioception token within the frame. Five of its layers are Mamba-2 and the fourth is bidirectional attention. This encoder carries no state across timesteps. Its last position is read out as the frame feature $z_t\in\mathbb{R}^{768}$.

\paragraph{Recurrent Memory}

The recurrent memory summarizes the whole observation history. It is a stack of six blocks of width 768. Five blocks are Mamba-2 layers and have no feed-forward network (FFN). The fourth block is a causal attention layer and has one FFN. At every step the memory takes $z_t$ as input, updates its retained state, and returns a readout,

\begin{equation}
    (q_t,m_t)=R_\phi(z_t,q_{t-1}),
    \label{eq:memory}
\end{equation}
where $q_t=(h_t,C_t)$. Here $h_t$ denotes the persistent recurrent states of the five Mamba-2 layers, and $C_t$ is the key-value cache of the attention layer. The size of $h_t$ is fixed by the architecture. The cache $C_t$ holds one entry per step and therefore grows linearly with episode length. Each entry is one 768-dimensional vector, so the memory never stores images or patch tokens. All of $q_t$ is reset at the start of each episode. The readout $m_t$ is the normalized output of the last block, not the state itself.

Two properties of the memory are set by configuration. The first is capacity: the number of blocks and the state width $d_{\mathrm{state}}$ of each Mamba-2 layer. The second is the update rule of the recurrent layers. Mamba-2 is the default. For comparison we replace it with Mamba-3 or Gated DeltaNet-2 and keep the observation encoder, the attention layer, the action decoder, and the training recipe unchanged. We match parameter counts across update rules. Table~\ref{tab:updaterules} summarizes what each rule does to information already in the state.
During training, the recurrence runs over the whole trajectory as a parallel scan. At inference, it runs one step at a time.

\begin{table}[t]
\centering
\caption{Update rules of the recurrent layers compared in Sec.~\ref{sec:eval}. Each step writes a cue-content pair $(k_t,v_t)$ derived from $z_t$ into the state $h_t$. The rules differ in what the write does to information already stored. The descriptions give the intended semantics of each rule. Whether a trained policy uses them this way is an empirical question.}

\label{tab:updaterules}

\scriptsize
\setlength{\tabcolsep}{2.5pt}
\renewcommand{\arraystretch}{1.12}
\begin{tabularx}{\columnwidth}{
    @{}p{1.15cm}
    p{2.75cm}
    >{\raggedright\arraybackslash}X@{}
}
\toprule
\textbf{Model} & \textbf{State update} & \textbf{Update semantics} \\
\midrule
Mamba-2 &
$\mathrm{diag}(a_t)h_{t-1}+v_tk_t^\top$ &
\textbf{Accumulate:} storing a similar cue again adds another write to its
trace, so repeated occurrences of an event build up in the
state, the property counting requires. \\
\addlinespace
Mamba-3 &
$\mathrm{diag}(a_te^{i\theta_t})h_{t-1}+v_tk_t^\top$ &
\textbf{Accumulate + rotate:} as above, with phase evolution that provides
an additional signal for temporal progression or motion. \\
\addlinespace
GDN-2 &
$\alpha_t(I-\beta_tk_tk_t^\top)h_{t-1}+\beta_tv_tk_t^\top$ &
\textbf{Replace:} storing a similar cue overwrites its previous content
with the latest value, while other stored cues remain undisturbed.
Repeated identical events therefore leave a single trace. \\
\bottomrule
\end{tabularx}

\end{table}

\paragraph{Observation and memory conditioned Action Generation}

Following flow-matching based policies~\cite{reuss2025flowerdemocratizinggeneralistrobot, vanjani2025disdp, reuss2023goal}, a Transformer decoder predicts an action chunk $A_t\in\mathbb{R}^{K\times d_a}$ with $K=16$. For Gaussian noise $X_0$ and $\tau\sim\mathcal{U}[0,1]$, we set $X_\tau=(1-\tau)X_0+\tau A_t$ and minimize

\begin{equation}
    \mathcal{L}_{\mathrm{FM}}
    =
    \mathbb{E}\!\left[
        \left\|
        v_\theta(X_\tau,\tau;z_t,m_t)-(A_t-X_0)
        \right\|_2^2
    \right].
\end{equation}

The decoder has four blocks. Each block applies self-attention over the action tokens, then cross-attention to $z_t$, then cross-attention to $m_t$, then a feed-forward network. The two cross-attention operations have separate parameters, so the current observation and the memory enter every block through separate paths.  Flow time $\tau$ conditions every block through zero-initialized adaptive layer normalization.

\paragraph{Implementation and training}

Training processes each demonstration as one causal scan over the full episode. The vision and text backbones stay frozen and are adapted with LoRA of rank 16 and 8, respectively. Real-robot policies train for 500 epochs. At inference, the decoder integrates the velocity field with $N=4$ Euler steps. The robot executes the first $k=4$ actions of each chunk and then replans. The frame encoder runs and the memory advances at every control step, including during chunk execution. The memory's recurrent states and attention cache are reset between episodes.

\section{Evaluation}\label{sec:eval}
We evaluate \smriti on two simulation benchmarks and three real-robot tasks. The evaluation asks three questions:

\begin{itemize}
\item \textbf{(RQ1)} How effective is \smriti on standard and on memory-dependent manipulation?
\item \textbf{(RQ2)} How do its observation encoder and its memory conditioning affect performance?
\item \textbf{(RQ3)} How do the recurrent update rule and the memory capacity affect different memory demands?
\end{itemize}
We report stage-wise behavior within these sections and computational cost separately.

\subsection{Experimental setup}

\begin{figure}[t]
\centering

\setlength{\tabcolsep}{1.2pt}
\renewcommand{\arraystretch}{1.0}

\newcommand{\taskw}{0.035\textwidth}

\newcommand{\stepw}{0.147\columnwidth}
\newcommand{\steph}{0.215\columnwidth}

\definecolor{aliascol}{RGB}{204,85,0}

\newcommand{\stepimg}[3][0 0 0 84]{%
\begin{tikzpicture}[baseline=(img.center)]
    \node[inner sep=0] (img) {%
        \includegraphics[
            width=\stepw,
            height=\steph,
            trim=#1, clip
        ]{#3}%
    };
    \node[
        anchor=south,
        minimum width=\stepw,
        text width=\stepw,
        align=center,
        fill=white,
        fill opacity=0.78,
        text opacity=1,
        inner xsep=0pt,
        inner ysep=1.5pt,
        minimum height=2.6ex,
        font=\footnotesize
    ] at (img.south) {#2};
\end{tikzpicture}%
}

\newcommand{\stepimgA}[3][0 0 0 84]{%
\begin{tikzpicture}[baseline=(img.center)]
    \node[inner sep=0] (img) {%
        \includegraphics[
            width=\stepw,
            height=\steph,
            trim=#1, clip
        ]{#3}%
    };
    \node[
        anchor=south,
        minimum width=\stepw,
        text width=\stepw,
        align=center,
        fill=white,
        fill opacity=0.78,
        text opacity=1,
        inner xsep=0pt,
        inner ysep=1.5pt,
        minimum height=2.6ex,
        font=\footnotesize
    ] at (img.south) {#2};
    \draw[aliascol, line width=1.2pt, rounded corners=1pt]
        ($(img.south west)+(0.7pt,0.7pt)$) rectangle ($(img.north east)-(0.7pt,0.7pt)$);
\end{tikzpicture}%
}

\newcommand{\aliasnote}[1]{%
    \multicolumn{6}{c}{%
        \parbox{\dimexpr 0.882\columnwidth+10\tabcolsep\relax}{\centering\scriptsize\color{aliascol}%
        \tikz[baseline=-0.6ex]\draw[aliascol, line width=1.1pt] (0,0) rectangle (0.75em,0.62em);%
        \; #1}%
    }%
}

\newcommand{\tasklab}[1]{%
    \raisebox{-0.5\height}{%
        \rotatebox[origin=c]{90}{%
            \normalsize\bfseries\textsc{#1}%
        }%
    }%
}

\begin{tabular}{@{}c@{\hspace{3pt}}cccccc@{}}

\tasklab{plant}
&
\stepimg[0 84 0 0]{Initial scene}{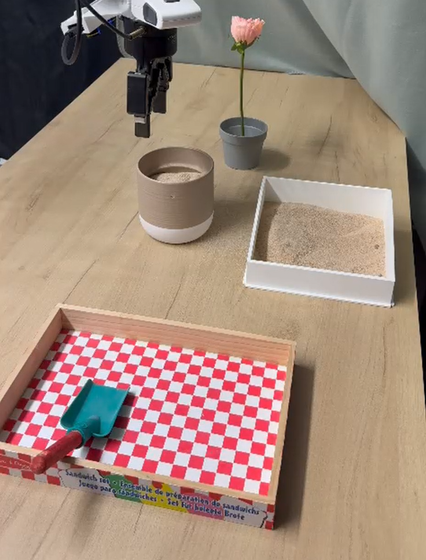}
&
\stepimg[0 84 0 0]{Pick up shovel}{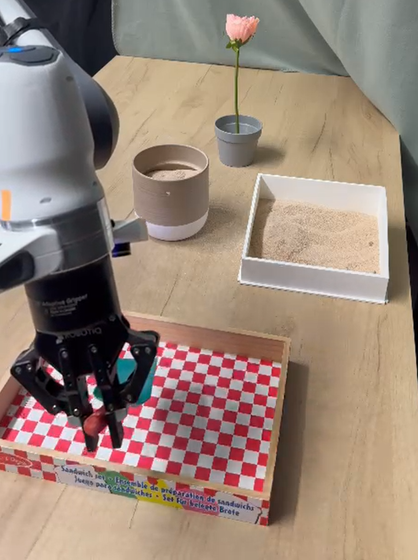}
&
\stepimgA[0 84 0 0]{First scoop}{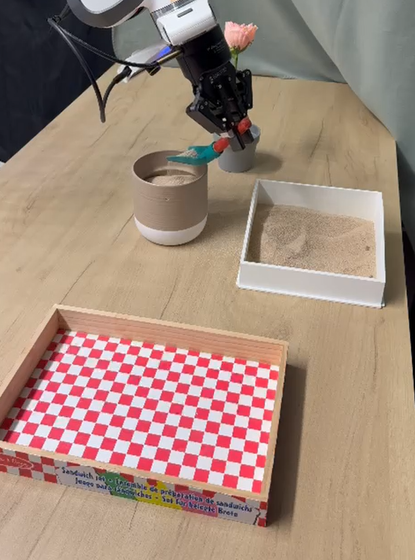}
&
\stepimgA[0 84 0 0]{Second scoop}{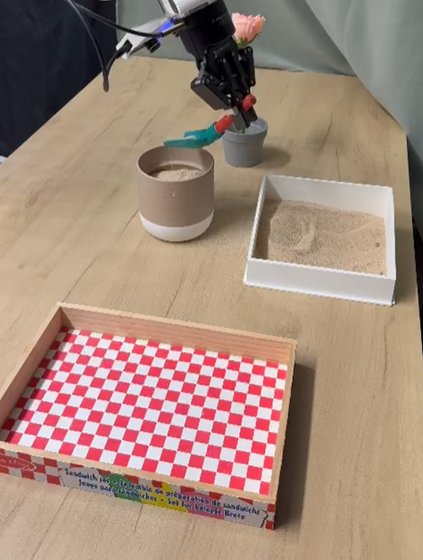}
&
\stepimg[0 84 0 0]{Put shovel down}{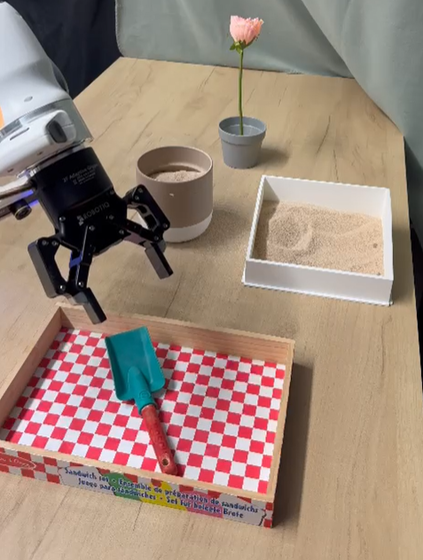}
&
\stepimg[0 84 0 0]{Put flower in the pot}{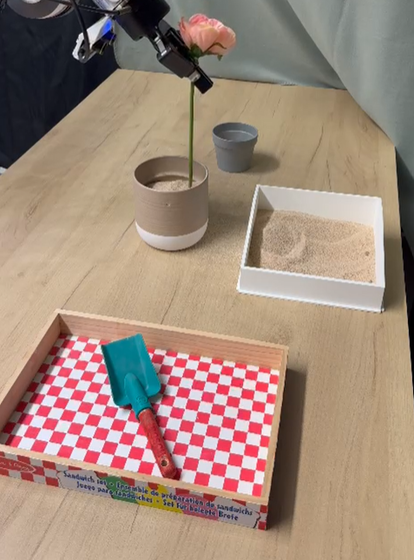}
\\[1pt]
& \aliasnote{same view after either scoop: \emph{scoop again} vs.\ \emph{put shovel down}, decided by the count}
\\[4pt]

\tasklab{pot timer}
&
\stepimg{Initial scene}{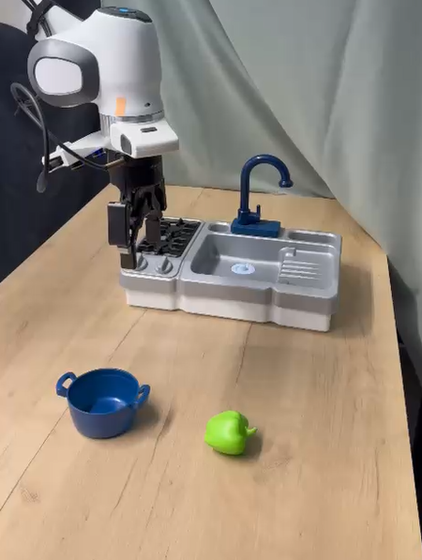}
&
\stepimg{Pick up pot}{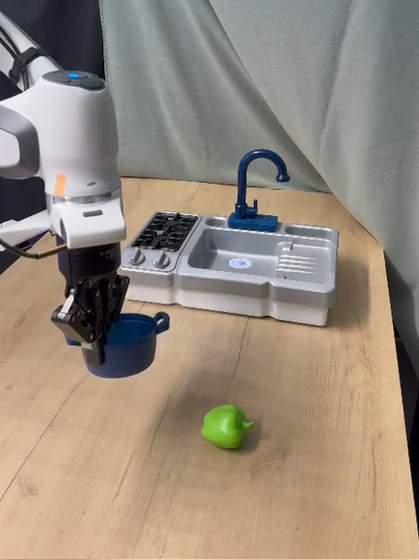}
&
\stepimgA{Put pot on stove}{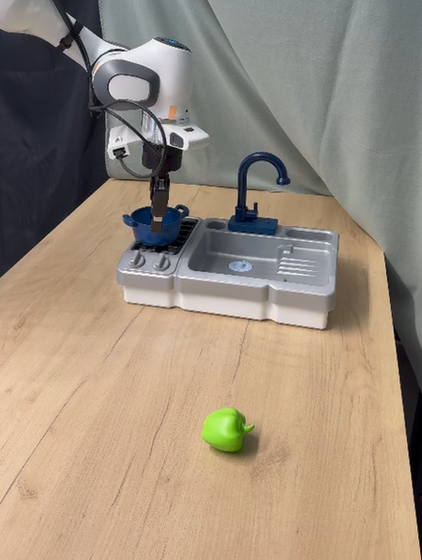}
&
\stepimgA{Wait 30\,s}{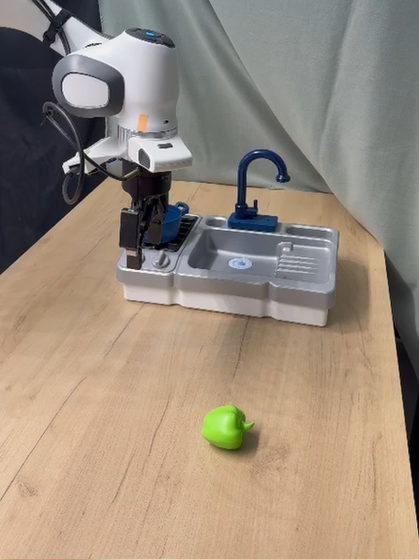}
&
\stepimg{Pick up pepper}{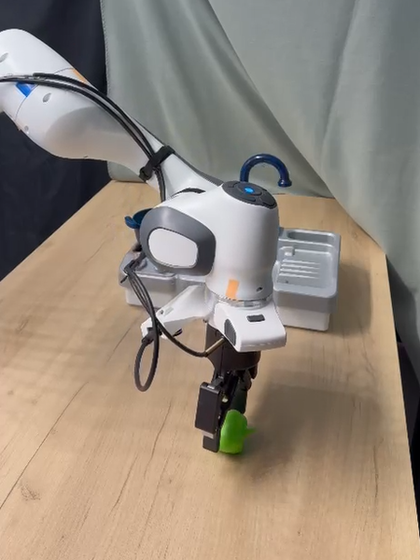}
&
\stepimg{Put pepper into pot}{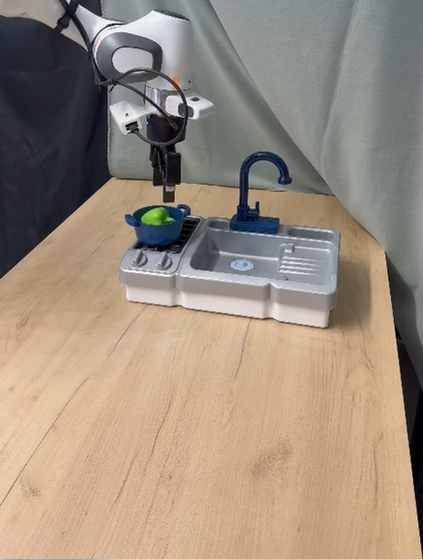}
\\[1pt]
& \aliasnote{identical frames throughout the wait: \emph{keep waiting} vs.\ \emph{add pepper}, decided by elapsed time}
\\[4pt]

\tasklab{sponge}
&
\stepimgA{Initial scene}{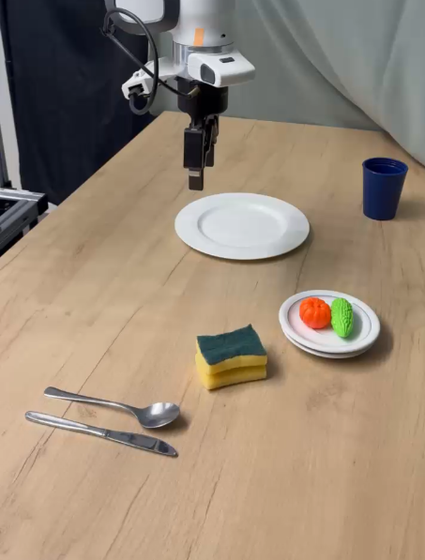}
&
\stepimg{Pick up sponge}{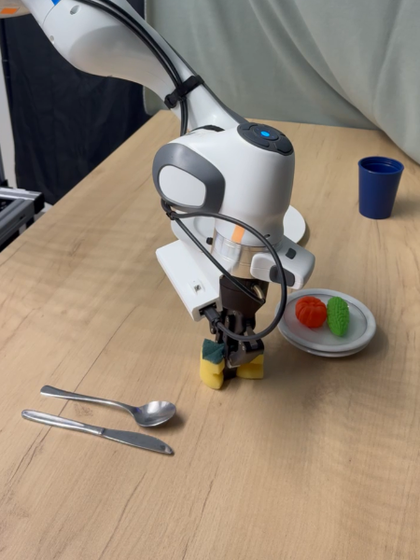}
&
\stepimg{Put sponge on plate}{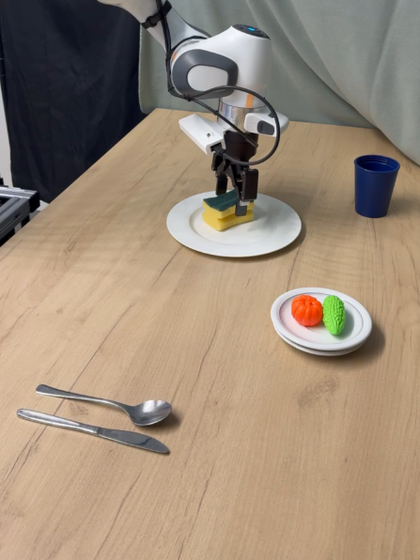}
&
\stepimg{Open gripper}{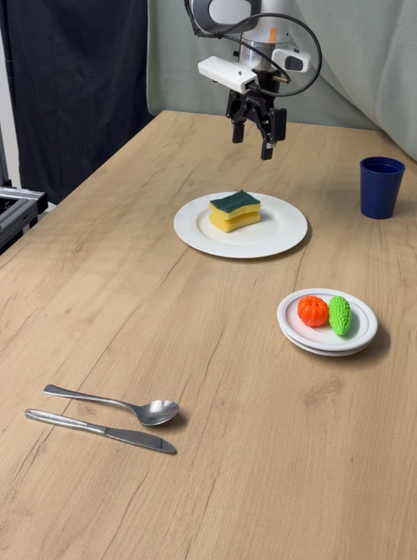}
&
\stepimgA{Pick sponge again}{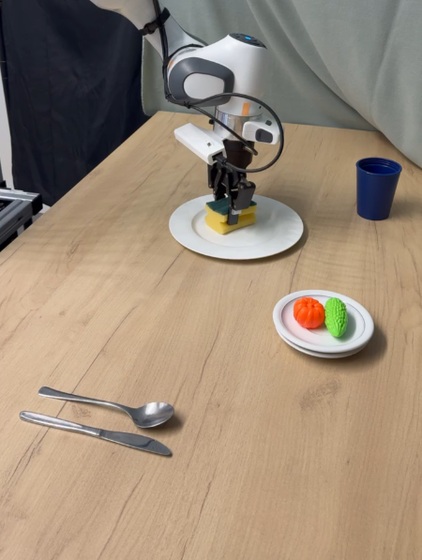}
&
\stepimg{Put it back to start position}{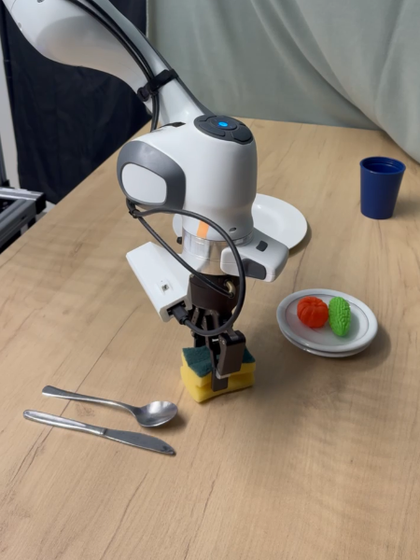}
\\[1pt]
& \aliasnote{the start position (left) has left the view by the re-grasp (right): the return target must be recalled}
\\

\end{tabular}

\caption{Real-robot tasks, one per row, each isolating one memory demand: \textsc{plant} counts scoops across five stages, \textsc{pot timer} times an interval while the scene is static, and \textsc{sponge} recalls an unmarked start position that has left the camera view. Outlined frames mark the decision point where the correct action depends on an earlier event, not on the current observation.}
\label{fig:real_tasks}

\end{figure}
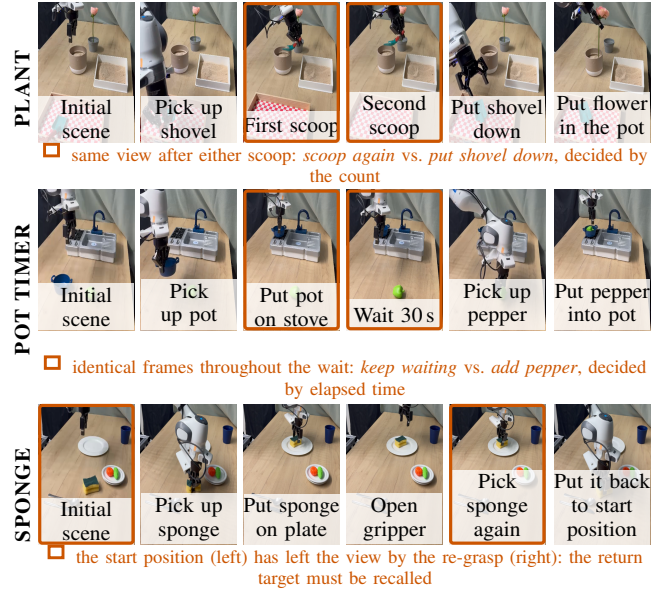

\paragraph{\textbf{Simulation benchmarks}}
We evaluate \smriti on LIBERO~\cite{liu2023libero}, including Spatial, Object, Goal, and LIBERO-10, using two cameras and 50 epochs. Each 10-task suite is evaluated over 500 episodes (50 per task). Since LIBERO is largely Markovian, it tests whether full-history recurrence preserves performance when the current observation is sufficient. We use LIBERO-Plus~\cite{fei2025libero} to evaluate robustness to controlled visual and physical perturbations. RMBench~\cite{chen2026rmbench} contains nine long-horizon bimanual tasks: five $M(1)$ and four $M(n)$ in which actions depend on past events no longer observable. Following its protocol, we train one policy per task on 50 demonstrations and evaluate 100 rollouts with unseen scenes.


\paragraph{\textbf{Real robot setup}}
We further evaluate the policies on three real-robot tasks that probe distinct memory demands under real-world execution variability. Unlike simulation, real-world manipulation introduces visual clutter, contact variation, friction, and trajectory variability, which can interfere with retaining task-relevant history. 
We therefore consider three tasks requiring \emph{spatial recall}, \emph{discrete event counting}, and \emph{interval timing}, each containing states where the correct action cannot be determined from the current observation alone. The three tasks are shown in Fig.~\ref{fig:real_tasks}. We use a single-arm Franka Emika Panda in the DROID setup~\cite{khazatsky2024droid}. The robot receives $224\times224$ RGB images from a fixed right-side camera and a wrist-mounted camera, together with joint-space proprioception. Demonstrations are collected through teleoperation at 15\,Hz. The policy predicts an eight-dimensional action comprising seven joint positions and a continuous gripper width.

The \textbf{\textsc{Sponge}} task evaluates spatial recall under clutter and broad spatial variation. The sponge is initialized across the robot's reachable workspace without any positional markers indicating its start location. The robot places it on a plate, releases it, and must then return it to its original position, which is no longer directly observable once the sponge has moved. This requires the remembered location to be maintained through the intervening manipulation and surrounding clutter. 48 demonstrations, $\sim$480 steps per episode.


The \textbf{\textsc{Plant}} task evaluates discrete event counting within a long-horizon, multi-stage manipulation sequence. The robot must place exactly two scoops of soil into a pot, put down the shovel, and pick up and insert a flower into the pot. Since the observations during the two scooping cycles are visually similar, the policy must track the number of completed scoops to determine whether to repeat the scooping action or proceed to the remaining stages. 45 demonstrations, $\sim$1{,}200 steps per episode, with varying scoop durations ($125.4 \pm 24.3$ frames).

The \textbf{\textsc{Pot Timer}} task evaluates interval timing. The robot places a pot on a stove, waits for approximately 30 seconds, and then picks up a pepper and places it in the pot. The visual scene remains nearly unchanged during the waiting period and provides no direct indication of elapsed time, requiring the policy to maintain temporal state internally. 45 demonstrations, $\sim$980 steps per episode.

\paragraph{Protocol}
We train one policy per task and configuration with a single seed for 500 epochs and evaluate the final checkpoint. Each policy runs 20 rollouts per task with randomized initial object positions. We report final success and the stage each rollout reached. In \textsc{Sponge}, a return counts as success when the sponge lands within 2-4\,cm of its original centre. In \textsc{Pot Timer}, a wait counts as correct when it lasts 29-31\,s. With 20 rollouts, one rollout is five percentage points, and differences of that size are within rollout noise.


\paragraph{Metrics}
We report \emph{success rate} in all evaluations. For the real robot we also report \emph{policy inference latency} and its reciprocal rate (Table~\ref{tab:compute}), and \emph{trajectory smoothness} as the average spectral arc length (SPARC)~\cite{scholp2021spectral} of the joint velocity profiles across all joints. Higher SPARC is smoother.

\subsection{Baselines}
For LIBERO and RMBench we report results from prior work. For the real-robot tasks we retrained every baseline on our demonstrations with the same cameras, control rate, action space, and rollout protocol. The short-history variants are our own extensions, since neither base method ships one. The baselines cover three ways a policy can access the past.
All baselines run on the robot's inference GPU 4060Ti. This excludes memory-augmented VLAs built on multi-billion-parameter backbones~\cite{shi2026memoryvla}. An official implementation of DSSP~\cite{guan2026dssp} was not available at the time of our experiments, so we discuss it as concurrent work and restrict comparisons to the methods above and to reported benchmark results.
 The baselines cover three ways a policy can access the past.
\textbf{(i) No memory:} \textbf{DP} is a standard Diffusion Policy
conditioned on the current right- and wrist-view images and proprioception, and \textbf{X-VLA}~\cite{zheng2026x} is a vision-language-action policy with no persistent temporal memory.
\textbf{(ii) Short-term memory:} \textbf{DP-H} and \textbf{X-VLA-H} add the last eight observations (stride 10), encoded by a shared trainable ResNet-18 and by a frozen Florence-2 encoder compressing them into eight history tokens through a two-layer Transformer, respectively. \textbf{(iii) Explicit attention-based memory:} \textbf{Gated Memory Policy (GMP)}~\cite{gao2026gated} retrieves visual and action information from a finite stored history using cross-attention and a calibrated binary memory gate.
Diffusion Policy is the standard imitation-learning policy at a size comparable to ours (268M parameters). X-VLA is a 0.9B-parameter pretrained VLA and one of the strongest model that fits the robot's GPU. GMP is the closest published memory policy at comparable size. All baselines were selected to be deployable on the robot's inference GPU, which excludes recent memory-augmented VLAs built on multi-billion-parameter backbones~\cite{shi2026memoryvla}.


\subsection{Results and Discussion}
Unless stated otherwise, \smriti denotes the Mamba-2 instantiation.



\textbf{RQ1: \smriti is competitive on standard manipulation and strong on memory-dependent manipulation.}
On LIBERO, \smriti reaches 95.3\% average success across the four suites (Table~\ref{tab:libero_main}). Diffusion Policy reaches 72.4\%. X-VLA, which uses large-scale pretraining, reaches 98.1\%. On LIBERO-Plus, \smriti reaches 64.2\% and X-VLA 71.4\%. \smriti therefore remains competitive on tasks that the current observation alone can solve.

On RMBench, \smriti reaches 62.4\% average success over nine tasks (Table~\ref{tab:rmbench}). It has the best or tied-best score on six of them, including three of the four $M(n)$ tasks. EventVLA has a higher overall average of 67.9\%. It builds on a 4B-parameter backbone, about ten times larger than \smriti, with an additional large-scale pretraining phase. Performance varies across tasks. \smriti reaches 100\% on Put Back Block and Swap Blocks but 4\% on Observe and Pick Up and 6\% on Press Button. Chronos~\cite{zhou2026chronos} is not in the table because we could not establish a matched protocol as it is not evaluated on all tasks.

The real-robot tasks show the largest gap (Table~\ref{tab:steps}). Policies without persistent memory succeed in 0-8.3\% of rollouts. X-VLA-H, the strongest baseline, reaches 8.3\% (95\% Wilson CI [3.6\%, 18.1\%]). \smriti variants reach 50.0-66.7\%, and the best variant reaches 66.7\% (95\% CI [54.1\%, 77.3\%]). The two intervals do not overlap. All three \smriti variants wait the correct interval in every \textsc{Pot Timer} rollout, and the Mamba variants complete 85-90\% of \textsc{Plant} rollouts. The stage-wise columns of Table~\ref{tab:steps} show where the baselines fail, which we examine next.

\textit{Stage-wise view.} The three tasks share a structure: observation-driven stages lead to one decision that the current observation cannot resolve. Table~\ref{tab:steps} breaks each rollout down by stage. Baselines complete the early stages and fail at that decision. X-VLA-H places the sponge on the plate in 15 of 20 rollouts but returns it to the right place in only 2. In \textsc{Plant}, it scoops at least once in 19 of 20 rollouts but completes the task in 2, and in 17 of them it keeps scooping past two. In \textsc{Pot Timer}, it places the pot in every rollout but waits the correct interval in 1. \smriti completes the decision in most rollouts, and the failures that remain, occur before the temporal memory decision stage.

\begin{table}[t]
    \centering
    \caption{
        Success rates (\%) on LIBERO. 
    }
    \label{tab:libero_main}
    \begingroup
    \footnotesize
    \setlength{\tabcolsep}{4pt}
    \renewcommand{\arraystretch}{1.05}
    \resizebox{\ifdim\width>\columnwidth\columnwidth\else\width\fi}{!}{%
    \begin{tabular}{@{} l c c c c c c @{}}
        \toprule
        Method & Spatial & Object & Goal & LIBERO-10 & Avg. & LIBERO-Plus \\
        \midrule
        Diffusion policy  & 78.3 & 92.5 & 68.3 & 50.5 & 72.4 & \tbd \\
        X-VLA~\cite{zheng2026x} & \textbf{98.2} & \textbf{98.6} & \textbf{97.8} & \textbf{97.6} & \textbf{98.1} & \textbf{71.4} \\
        \rowcolor{oursblue}
        Ours              & 96.4 & 97.2 & 97.2 & 90.4 & 95.3 & 64.2 \\
        \bottomrule
    \end{tabular}%
    }
    \endgroup
\end{table}

\begin{table}[t]
    \centering
    \caption{
        Success rates (\%) on RMBench. * Indicates models utilizing large scale pretraining.
    }
    \label{tab:rmbench}
    \begingroup
    \setlength{\tabcolsep}{3pt}
    \renewcommand{\arraystretch}{1.05}
    \footnotesize
    \resizebox{\ifdim\width>\columnwidth\columnwidth\else\width\fi}{!}{%
        \begin{tabular}{
            @{} l c
            c c c c c c c c
            >{\columncolor{oursblue}}c @{}
        }
            \toprule
            Task & TMC
            & \shortstack{DP\\\textcolor{green!50!black}{(90M)}}
            & \shortstack{ACT\\\textcolor{green!50!black}{(80M)}}
            & \shortstack{$\/*pi_{0.5}$\\\textcolor{red!75!black}{(3B)}}
            & \shortstack{*X-VLA\\\textcolor{green!50!black}{(0.9B)}}
            & \shortstack{MEM-0\\\textcolor{red!75!black}{(10B)}}
            & \shortstack{EventVLA\\\textcolor{red!75!black}{(4B)}}
            & \shortstack{*MemoryVLA\\\textcolor{red!75!black}{(7.3B)}}
            & \shortstack{*MemER\\\textcolor{red!75!black}{(10B)}}
            & \shortstack{Ours\\\textcolor{green!50!black}{(374M)}} \\
            \midrule

            Observe and Pick Up & $M(1)$
            & 1 & 1 & 9 & 9
            & 4 & \textbf{21} & 2 & 7 & 4 \\

            Rearrange Blocks & $M(1)$
            & 0 & 29 & 13 & 13
            & 89 & \textbf{96} & 53 & 17 & \textbf{96} \\

            Put Back Block & $M(1)$
            & 0 & 0 & 11 & 18
            & 90 & 95 & 81 & 0 & \textbf{100} \\

            Swap Blocks & $M(1)$
            & 11 & 2 & 24 & 16
            & 67 & 96 & 76 & 14 & \textbf{100} \\

            Swap T & $M(1)$
            & 20 & 2 & 15 & 3
            & 14 & \textbf{87} & 9 & 7 & 64 \\

            \addlinespace[2pt]
            \textit{Average} & $M(1)$
            & 6.4 & 6.8 & 14.4 & 11.8
            & 52.8 & \textbf{79.0} & 44.2 & 9.0 & 72.8 \\

            \midrule

            Battery Try & $M(n)$
            & 10 & 19 & 16 & 26
            & 28 & 35 & 33 & 27 & \textbf{43} \\

            Blocks Ranking Try & $M(n)$
            & 10 & 0 & 6 & 1
            & 18 & 81 & 53 & 12 & \textbf{99} \\

            Cover Blocks & $M(n)$
            & 0 & 0 & 0 & 2
            & 68 & \textbf{97} & 69 & 40 & 50 \\

            Press Button & $M(n)$
            & 0 & 0 & 0 & 0
            & 0 & 3 & 0 & 0 & \textbf{6} \\

            \addlinespace[2pt]
            \textit{Average} & $M(n)$
            & 5.0 & 4.8 & 5.5 & 7.3
            & 28.5 & \textbf{54.0} & 38.8 & 19.8 & 49.5 \\

            \midrule

            \textbf{Overall average} & --
            & 5.8 & 5.9 & 10.4 & 9.8
            & 42.0 & \textbf{67.9} & 41.8 & 13.8 & 62.4 \\

            \bottomrule
        \end{tabular}%
    }
    \endgroup
\end{table}
%
%
\definecolor{succ1}{RGB}{236,247,239}
\definecolor{succ2}{RGB}{214,238,221}
\definecolor{succ3}{RGB}{188,226,199}
\definecolor{succ4}{RGB}{160,212,177}
\definecolor{faillight}{RGB}{251,236,236}
\definecolor{avggray}{RGB}{232,234,238}
\begin{table*}[t]
\centering\footnotesize\setlength{\tabcolsep}{3.5pt}
\renewcommand{\arraystretch}{1.05}
\caption{Real-robot results, step by step. Each task moves from observation-driven stages to a later decision that depends on retained history (outlined in Fig.~\ref{fig:real_tasks}). Breaking results down this way, instead of reporting only final success, shows exactly where a policy fails. For \textsc{Plant}, the three scoop columns are mutually exclusive: a rollout is counted under the number of scoops it performed before it either moved on to the flower or stayed in the scooping loop until timeout. For \textsc{Pot Timer}, ``No wait'' and ``Wrong wait'' are rollouts that left the waiting state early or outside the 29-31\,s window. For \textsc{Sponge}, ``Returned to wrong position'' and ``Returned to same position'' partition the rollouts that placed the sponge, except rollouts that stalled at the plate. Success is the final column of each task. All values are percentages of 20 rollouts with randomised start positions. Avg. is the mean over the success rate of three tasks.}
\label{tab:steps}\label{tab:real_main}
\resizebox{\ifdim\width>\textwidth\textwidth\else\width\fi}{!}{%
\begin{tabular}{l |
  >{\columncolor{faillight}}c >{\columncolor{succ3}}c >{\columncolor{faillight}}c >{\columncolor{succ4}}c |
  >{\columncolor{succ1}}c >{\columncolor{faillight}}c >{\columncolor{faillight}}c >{\columncolor{succ2}}c >{\columncolor{succ3}}c >{\columncolor{succ4}}c |
  >{\columncolor{succ1}}c >{\columncolor{faillight}}c >{\columncolor{succ4}}c |
  c}
\toprule
 & \multicolumn{4}{c|}{Plant} & \multicolumn{6}{c|}{Pot timer} & \multicolumn{3}{c|}{Sponge} & \\
\cmidrule(lr){2-5}\cmidrule(lr){6-11}\cmidrule(lr){12-14}
Method
  & 1 scoop & 2 scoops & 3 or more & \shortstack{Flower planted\\Success}
  & \shortstack{Pot on\\stove} & No wait & Wrong wait & \shortstack{$\approx$30\,s\\wait} & \shortstack{Pepper in\\the pot} & Success
  & \shortstack{Pick sponge,\\put on plate} & \shortstack{Returned to\\wrong position} & \shortstack{Returned to\\same position\\Success}
  & Avg.\\
\midrule
\rowcolor{liberogray}
\multicolumn{15}{l}{\textbf{Baselines}}\\
Diffusion policy~\cite{chi2023diffusion}               & -- & -- & -- & 0 & -- & -- & -- & -- & -- & 0 & -- & -- & 0 & 0.0\\
Diffusion policy~\cite{chi2023diffusion} + history     & 15 & 10 & 0 & 0 & 100 & 25 & 0 & 0 & 25 & 0 & 15 & -- & 0 & 0.0\\
X-VLA~\cite{zheng2026x}                                & -- & -- & -- & 0 & -- & -- & -- & -- & -- & 0 & -- & -- & 0 & 0.0\\
X-VLA~\cite{zheng2026x} + history                      & 0 & 10 & 85 & 10 & 100 & 0 & 35 & 5 & 40 & 5 & 75 & 65 & 10 & 8.3\\
Gated Memory Policy (GMP)~\cite{gao2026gated}          & 65 & 20 & 10 & 15 & 95 & 20 & 10 & 5 & 35 & 5 & 50 & 50 & 0 & 6.7\\
\midrule
\rowcolor{liberogray}
\multicolumn{15}{l}{\textbf{Ours}}\\
\textbf{\smriti-Mamba-2}            & 0 & 100 & 0 & 85 & 100 & 0 & 0 & 100 & 95 & 95 & 20 & 0 & 20 & \cellcolor{avggray}\textbf{66.7}\\
\textbf{\smriti-Mamba-3}            & 0 & 95 & 0 & \textbf{90} & 100 & 0 & 0 & 100 & 100 & \textbf{100} & 0 & 0 & 0 & \cellcolor{avggray}63.3\\
\textbf{\smriti-Gated DeltaNet 2}   & 0 & 70 & 0 & 20 & 100 & 0 & 0 & 100 & 90 & 90 & 40 & 0 & \textbf{40} & \cellcolor{avggray}50.0\\
\bottomrule
\end{tabular}%
}
\end{table*}

%
\begin{table*}[h]
\centering\footnotesize\setlength{\tabcolsep}{3.5pt}
\renewcommand{\arraystretch}{1.05}
\caption{Step-wise real-robot success across six memory-to-action conditioning mechanisms. Columns as in Table~\ref{tab:steps}.}
\label{tab:mem_steps}\label{tab:mem_integration}\label{fig:integration_progress}
\resizebox{\ifdim\width>\textwidth\textwidth\else\width\fi}{!}{%
\begin{tabular}{l |
  >{\columncolor{faillight}}c >{\columncolor{succ3}}c >{\columncolor{faillight}}c >{\columncolor{succ4}}c |
  >{\columncolor{succ1}}c >{\columncolor{faillight}}c >{\columncolor{faillight}}c >{\columncolor{succ2}}c >{\columncolor{succ3}}c >{\columncolor{succ4}}c |
  >{\columncolor{succ1}}c >{\columncolor{faillight}}c >{\columncolor{succ4}}c |
  >{\columncolor{avggray}}c}
\toprule
 & \multicolumn{4}{c|}{Plant} & \multicolumn{6}{c|}{Pot timer} & \multicolumn{3}{c|}{Sponge} & \\
\cmidrule(lr){2-5}\cmidrule(lr){6-11}\cmidrule(lr){12-14}
Integration
  & 1 scoop & 2 scoops & 3 or more & \shortstack{Flower planted\\Success}
  & \shortstack{Pot on\\stove} & No wait & Wrong wait & \shortstack{$\approx$30\,s\\wait} & \shortstack{Pepper in\\the pot} & Success
  & \shortstack{Pick sponge,\\put on plate} & \shortstack{Returned to\\wrong position} & \shortstack{Returned to\\same position}
  & Avg.\\
\midrule
\rowcolor{liberogray}

\textbf{\smriti, cross-attention} & 0 & 100 & 0 & \textbf{85} & 100 & 0 & 0 & 100 & 95 & \textbf{95} & 20 & 0 & 20 & \textbf{66.7}\\
\midrule
\rowcolor{liberogray}
\multicolumn{15}{l}{\textbf{Integration variants}}\\
Late fusion   & 0 & 0 & 0 & 0 & 90 & 0 & 0 & 90 & 90 & 90 & 70$^\dagger$ & 40 & \textbf{25} & 38.3\\
AdaLN         & 35 & 15 & 0 & 5 & 100 & 0 & 0 & 100 & 90 & 90 & 25 & 5 & 0 & 31.7\\
Scale         & 0 & 25 & 0 & 10 & 80 & 0 & 0 & 80 & 75 & 75 & 40 & 40 & 0 & 28.3\\
Dropout       & 5 & 10 & 0 & 0 & 90 & 0 & 0 & 90 & 60 & 60 & 20 & 10 & 0 & 20.0\\
Gate          & 35 & 5 & 0 & 5 & 50 & 0 & 0 & 50 & 20 & 20 & 20 & 0 & 0 & 8.3\\
\bottomrule
\end{tabular}%
}
\end{table*}

\textbf{RQ2: Both the observation encoder and the memory conditioning matter, and the largest effects appear on \textsc{Plant}.}

\textit{Observation encoder.} We fix the memory to Mamba-2 and vary the frame encoder (Table~\ref{tab:rq1}). The hybrid Mamba-attention encoder reaches 66.7\% average success. A full-Transformer encoder reaches 25.0\%. It drops from 85\% to 0\% on \textsc{Plant}, from 95\% to 65\% on \textsc{Pot Timer}, and from 20\% to 10\% on \textsc{Sponge}. A pure-SSM configuration removes the attention layer from both the frame encoder and the memory. It reaches 25\% on \textsc{Plant}, the only task on which we evaluated it. Removing the memory module gives 0\% on all three tasks. Removing the multimodal encoder also gives 0\%. The encoder supplies both the decoder input and the memory input, so these differences are end-to-end effects of representation quality. Counting is the most sensitive demand: repeated scoops must produce consistent representations before the memory can accumulate them. The no-memory result shows that the temporal module as a whole is necessary. It does not separate the contributions of its recurrent layers and its attention layer.
\begin{table}[h]
    \centering
    \caption{Real-robot success rates for observation-encoder and component ablations.}
    \label{tab:rq1}
    \begingroup
    \footnotesize
    \setlength{\tabcolsep}{3pt}
    \renewcommand{\arraystretch}{1.0}
    \resizebox{\ifdim\width>\columnwidth\columnwidth\else\width\fi}{!}{%
    \begin{tabular}{@{} l c c c >{\columncolor{oursblue}}c @{}}
        \toprule
        Setting & Sponge & Plant & Pot timer & Avg. \\
        \midrule
        Ours, hybrid encoder                         & \textbf{20} & \textbf{85} & \textbf{95} & \textbf{66.7} \\
        Ours, full-transformer encoder               & 10 & 0 & 65 & 25.0 \\
        Ours, pure-SSM encoder                       & \tbd & 25 & \tbd & \tbd \\
        Ours, w/o multimodal observation encoder tokens & 0 & 0 & 0 & 0.0 \\
        Ours, without memory module                  & 0 & 0 & 0 & 0.0 \\
        \bottomrule
    \end{tabular}%
    }
    \endgroup
\end{table}

\textit{Memory conditioning.} We fix the encoder and memory and vary how $m_t$ enters the four-block decoder (Table~\ref{tab:mem_steps}). Memory cross-attention in every block reaches 66.7\% average success. Cross-attention in the final two blocks only (late fusion) reaches 38.3\%. Modulation through AdaLN, a learned scale, or a content gate reaches 31.7\%, 28.3\%, and 8.3\%. Memory dropout changes training rather than the readout and reaches 20.0\%. The largest gap is on \textsc{Plant}, where every-block cross-attention reaches 85\% and no alternative exceeds 10\%. Several alternatives fail before the count matters. Late fusion never picks up the shovel. Dropout and Scale miss it in 17 and 15 of 20 rollouts. The Gate and AdaLN variants stop after a single scoop in 7 of 20 rollouts each. Weak conditioning therefore harms both manipulation and the count readout. In these evaluations, \textsc{Pot Timer} is less sensitive. Every variant that places the pot also waits the correct interval. Late fusion and AdaLN reach 90\%, against 95\% for every-block cross-attention. The Gate variant places the pot in only 10 of 20 rollouts, and in 6 of the other 10 it reaches for the pepper first. \textsc{Sponge} is the exception. Late fusion reaches the plate in 14 of 20 rollouts and returns the sponge correctly in 5 of them, for the highest observed success at 25\%. Every-block cross-attention reaches the plate in 4 of 20 and returns all 4, for 20\%. That difference is one rollout. No evaluated variant reaches high \textsc{Sponge} success. Grasping and spatial recall must both be learned from 48 demonstrations with broad position variation, and conditioning choices do not close that gap.


\begin{table}[h]
    \centering
\caption{Computational cost on the same hardware and inference setting at. Latency is per policy inference; the rate is its reciprocal. SPARC: higher is smoother.}
    \label{tab:compute}
    \begingroup
    \footnotesize
    \setlength{\tabcolsep}{4pt}
    \renewcommand{\arraystretch}{1.05}
    \resizebox{\ifdim\width>\columnwidth\columnwidth\else\width\fi}{!}{%
    \begin{tabular}{@{} l >{\columncolor{oursblue}}c c c c  @{}}
        \toprule
        Metric & \smriti & DP-H & X-VLA-H & GMP  \\
        \midrule
        Total / trainable parameters & 374M / 142M & 267.9M / 267.9M & 881.9M / 881.9M & 282.2M / 188.7M   \\

        Peak inference VRAM          & 1.44GB & 0.65GB & 2.55GB & 1.09GB  \\
        Inference latency/frequency & 0.059s/16.9hz & 0.11s/9hz & 0.288s/3.47hz & 0.093s/10.75hz  \\
        Smoothness metric (SPARC \cite{scholp2021spectral})    & -5.16 & -6.19 & -6.76 & -13.06  \\
        \bottomrule
    \end{tabular}%

    }
    \endgroup
\end{table}

\textbf{RQ3: The update rule and the memory capacity have task-dependent effects, and the additive rules have the highest observed success on counting.}

\textit{Update rule.} We keep the encoder, the attention layer, and the decoder fixed and replace the recurrent update (Table~\ref{tab:steps}). On \textsc{Plant}, Mamba-2 and Mamba-3 reach 85\% and 90\%. Gated DeltaNet-2 reaches 20\%. It records two scoops in 14 of 20 rollouts but completes the task in 4, so most of its failures come after the count is reached. On \textsc{Pot Timer}, all three rules wait the correct interval in every rollout and reach 90-100\% success. On \textsc{Sponge}, Gated DeltaNet-2 has the highest observed success at 40\%. Mamba-2 reaches 20\% and Mamba-3 0\%. The stage-wise columns qualify this gap. Mamba-2 and Gated DeltaNet-2 return the sponge correctly in every rollout that reaches the plate, 4 of 4 and 8 of 8. Mamba-3 never reaches the plate. The end-to-end difference on \textsc{Sponge} therefore arises before the recall step, in grasping and placement. The update rules motivate a hypothesis for the \textsc{Plant} pattern. Additive writes let evidence of repeated scoops accumulate, while the delta rule replaces the content stored under a matching cue and may represent multiplicity less directly (Table~\ref{tab:updaterules}). We do not measure the internal count, so this remains a hypothesis. Mamba-2 has the highest average success and is the default in all other experiments.

\textit{Memory capacity.} We fix the rule to Mamba-2 and vary depth and state width, with \textsc{Plant} as the most sensitive task. Reducing the memory from six blocks to two, which keeps the attention layer and leaves one Mamba-2 layer, lowers \textsc{Plant} success from 85\% to 65\%. For state width we evaluate $d_{\mathrm{state}}\in\{32,64,128,256\}$ with everything else fixed. \textsc{Plant} success is 15\%, 20\%, 85\%, and 0\%. At widths 32 and 64, 35-40\% of rollouts reach two scoops but do not complete the task due to precision failures. At width 256 the policy fails before reaching two scoops. Width 128 has the highest observed success in this sweep. The non-monotonic result shows sensitivity to the configuration under a single training run. It does not establish that larger states hold less. We use six blocks and width 128 in all other experiments.

\textbf{Computational efficiency.} Table~\ref{tab:compute} compares model size, inference cost, and trajectory smoothness under the same hardware and inference settings. \smriti has 374M parameters, of which 142M are trainable. Policy inference takes 59\,ms per step, against 110\,ms for DP-H, 93\,ms for GMP, and 288\,ms for X-VLA-H. The reciprocal rate is 16.9\,Hz. \smriti also has the highest SPARC, that is the smoothest end-effector trajectories, among the compared policies.

\FloatBarrier

\section{CONCLUSIONS}

We presented \smriti, a structured recurrent policy for memory-dependent manipulation. It's memory integrates the full observation history through Mamba-2 layers and one causal attention layer. It's flow-matching decoder reads the current observation and the memory through separate cross-attention in every block. This design reaches 95.3\% on LIBERO, 62.4\% on RMBench, and 66.7\% across three real-robot tasks that require spatial recall, counting, and timing. It also has the lowest measured inference latency among the compared memory-based policies, at 59\,ms per step.
The analysis explains where this performance comes from. Stage-wise results show that short-history policies execute the immediate manipulation but fail at the decision that depends on earlier information. Within the evaluated configurations, the hybrid observation encoder and memory cross-attention in every decoder block gave the highest average success. The recurrent update rule has task-dependent effects: the additive rules have the highest observed success on counting and timing, and the delta rule on spatial recall. Effective memory-dependent control therefore, depends on what is written to memory, how it is updated, and how the action generator reads it alongside the current observation.



\section{LIMITATIONS AND FUTURE WORK}

Our real-robot evaluation covers one task per memory demand, each evaluated over 20 rollouts. Broader studies across multiple tasks, embodiments, are needed to assess generalization, including multiple counts and durations distinguished by language instructions. The memory ablation removes the recurrent layers and the attention layer together, so their individual contributions remain open. The \textsc{Sponge} results show a remaining trade-off between memory use and fine-grained spatial control. Future work should examine these factors and improve success on spatial-return manipulation under limited demonstrations with larger task diversity.

\vspace{-2mm}
\section*{Acknowledgment}

We used Claude and ChatGPT to polish the text of this paper. The authors
reviewed every generated item, and all technical claims, experiments, and
results are our own. The research presented in this paper was
funded by the Deutsche Forschungsgemeinschaft (DFG, German Research Foundation) – 448648559. The authors gratefully acknowledge the computing time provided on the high-performance computer HoreKa by the National High-Performance Computing Center at KIT and by Gauss Centre for Supercomputing e.V. (www.gauss-centre.eu) for GCS Supercomputer JUPITER at Jülich Supercomputing Centre (JSC).

\bibliographystyle{IEEEtran}
\bibliography{references}

\end{document}